\documentclass[11pt,a4paper]{article}
\usepackage[hyperref]{eacl2021}
\usepackage{times}
\usepackage{latexsym}

\usepackage{amsmath}
\usepackage{booktabs}
\usepackage{multirow}
\usepackage{url}
\usepackage{xcolor}
\usepackage{subcaption}
\usepackage{graphicx}
\usepackage{lipsum}
\usepackage{minibox}

\usepackage{filecontents}
\usepackage{pgfplots}
\usepackage{pgfplotstable}
\pgfplotsset{compat=1.9}
\usetikzlibrary{arrows.meta,graphs,quotes,arrows,calc,positioning,patterns}

\definecolor{myred}{HTML}{d73027}
\definecolor{mylightred}{HTML}{fc8d59}
\definecolor{myyellow}{HTML}{fee090}
\definecolor{mylightblue}{HTML}{91bfdb}
\definecolor{myblue}{HTML}{4575b4}

\usepackage{array}
\newcolumntype{R}[1]{>{\raggedleft\let\newline\\\arraybackslash\hspace{0pt}}m{#1}}

\usepackage{microtype}

\aclfinalcopy 

\title{On User Interfaces for Large-Scale Document-Level\\Human Evaluation of Machine Translation Outputs}

\author{%
\minibox[c]{Roman\\Grundkiewicz} \qquad 
\minibox[c]{Marcin\\Junczys-Dowmunt} \qquad 
\minibox[c]{Christian\\Federmann} \qquad 
\minibox[c]{Tom\\Kocmi}\\[4mm]
  Microsoft, 1 Microsoft Way, Redmond, WA 98121, USA \\
  \texttt{\{Firstname.Lastname\}@microsoft.com} \\}
  
\date{}

\newsavebox{\measurebox}

\begin{document}
\maketitle
\begin{abstract}
Recent studies emphasize the need of document context in human evaluation of machine translations, but little research has been done on the impact of user interfaces on annotator productivity and the reliability of assessments. In this work, we compare human assessment data from the last two WMT evaluation campaigns collected via two different methods for document-level evaluation. Our analysis shows that a document-centric approach to evaluation where the annotator is presented with the entire document context on a screen leads to higher quality segment and document level assessments. It improves the correlation between segment and document scores and increases inter-annotator agreement for document scores but is considerably more time consuming for annotators.
\end{abstract}

\section{Introduction}

\begin{figure*}
\centering
  \begin{minipage}{.49\linewidth}
    \centering
    \subcaptionbox{The segment-level portion of the WMT19 interface.\label{fig:wmt19seg}}
      {\includegraphics[width=\linewidth]{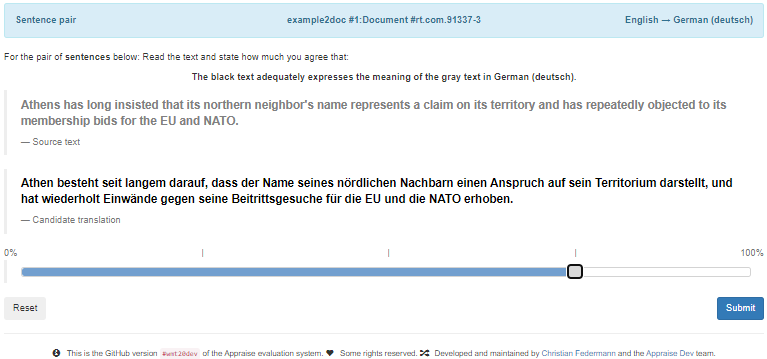}}\vspace{20pt}
    \subcaptionbox{The document-rating portion of the WMT19 interface.\label{fig:wmt19doc}}
      {\includegraphics[width=\linewidth]{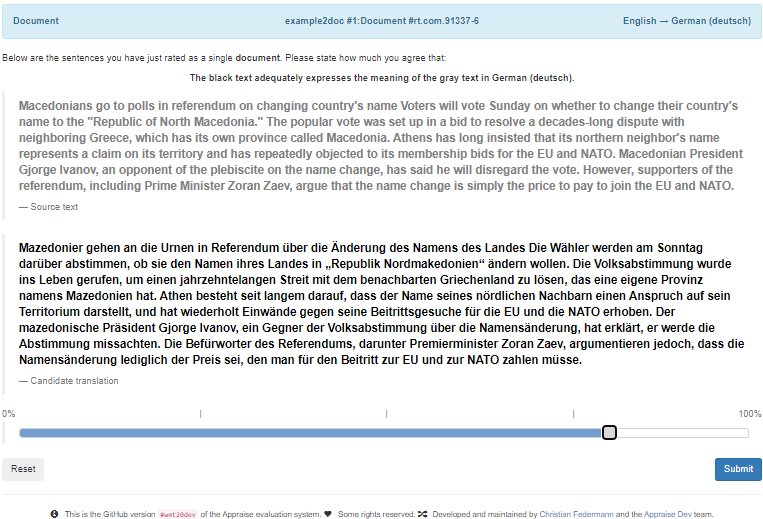}}
  \end{minipage}\qquad
  \begin{minipage}{.46\linewidth}
    \subcaptionbox{The document-centric WMT20 interface\label{fig:wmt20}}
      {\includegraphics[width=\linewidth]{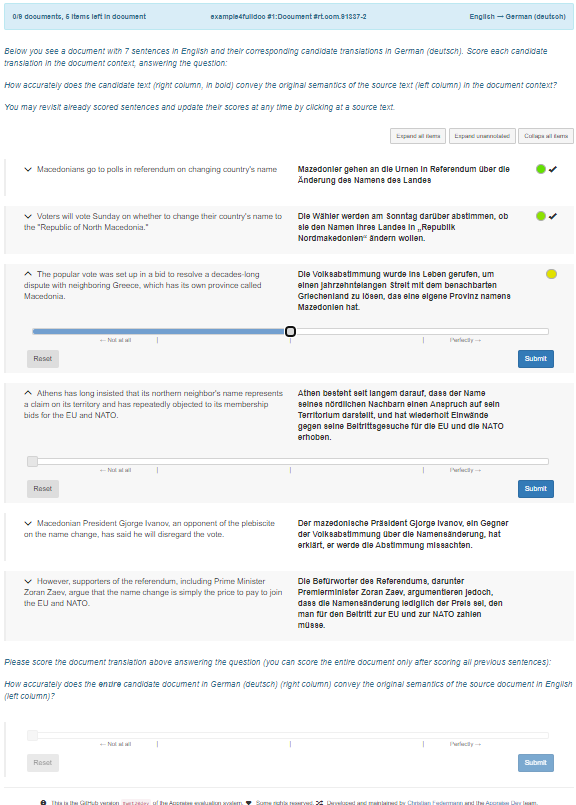}}
  \end{minipage}
\label{fig:appraise}
\caption{Screen shots of the Appraise interfaces used for the WMT19 (left) and WMT20 (right) human evaluation campaigns.\label{fig:wmt19}}
\end{figure*}

Recently, several studies have suggested that document context is required for the reliable human evaluation of machine-translated documents \cite{castilho-etal-2020-context,laubli2020jair}. With the improved performance of neural machine translation systems (NMT) over the past years, this is particularly important when assessing the potential for human parity or super-human performance of MT systems \cite{laubli-etal-2018-machine,toral-etal-2018-attaining}.
Following these recommendations, the WMT Conference on Machine Translation\footnote{\url{http://www.statmt.org/wmt20/}} has moved towards adopting and presenting document context in their human evaluation campaigns of 2019 and 2020 \cite{barrault-etal-2019-findings,barrault-etal-2020-findings}. The WMT campaigns are the largest academic efforts on human evaluation of machine-translated news articles in the field, running yearly since 2007.

At WMT19, the previous segment-level direct assessment evaluation \cite{bojar-etal-2017-findings,bojar-etal-2018-findings} --- where translated segments were presented to evaluators\footnote{In this work, we use the terms \emph{evaluator} and \emph{annotator} interchangeably.} in random order --- was extended by introducing ``segment ratings with document context'' \cite{barrault-etal-2019-findings}, and assessments of both, individual segments and entire documents, were collected. In this approach, segments from a single document translated by the same MT system were provided sequentially to evaluators in the order as they appear in the document, only one segment shown at a time (Fig.~\ref{fig:wmt19seg}), followed by the entire document comprised of already scored segments (Fig.~\ref{fig:wmt19doc}).
WMT 2020 \cite{barrault-etal-2020-findings} implemented a more document-centric approach, displaying the full translated document on a single screen (Fig.~\ref{fig:wmt20}) for most of the out-of-English language pairs.

While the change was primarily about the user interface (UI), we believe it can impact the quality of document-level evaluation to a large extent. \newcite{toral-2020-reassessing} has noticed potential issues arising from the limited inter-sentential context in the WMT19 method, in which the evaluator does not have continuous access to all segments from the document. Unable to revisit previous sentences and never seeing subsequent sentences, the evaluator might forget or lack access to important details necessary to rate the current segment. On the other hand, displaying a long document on a screen can notably increase cognitive load, potentially lowering reliability of assessments over time \cite{gonzalez2011cognitive}, and increase annotation time and costs, especially at the scale of the WMT evaluation campaigns.

In this work, we compare human assessment scores collected during the last two WMT evaluation campaigns and analyze the impacts of the user interface changes between these campaigns. We also attempt to determine whether switching to the document-centric UI was an improvement to the human evaluation procedure and should be adopted in future editions of WMT for all language pairs. We examine if and to what extent human raters make use of the document context, estimate the reliability of document ratings collected through both interfaces, and study potential additional costs resulting from the document-centric evaluation at a large scale.

\section{Document context in human evaluation of MT outputs}
\label{sec:context}

Recent research emphasized the importance of document context in human evaluation of machine translation, especially in terms of accessing potential human parity or super-human performance \cite{laubli-etal-2018-machine,toral-etal-2018-attaining,graham-etal-2020-statistical,toral-2020-reassessing}. 

Several works have compiled sets of recommendations for document-level evaluation. For example, \newcite{laubli2020jair} recommend evaluation of documents instead of independent sentences as translators tend to judge machine translation more favourably if they cannot identify errors related to textual coherence and cohesion due to lack of context.
\newcite{castilho-etal-2020-context} have examined the necessary context span needed for evaluation across different domains, and for relatively short documents like news articles, the authors recommend presenting the whole document during the assessment of individual segments.
Using document context has also been recommended by \newcite{toral-2020-reassessing} who reported that this information was needed for evaluators to rank systems in a contrastive evaluation setting.
Having the text available during the assessment of fluency or adequacy might be essential for some evaluators who spend more time reading than assessing \cite{castilho-2020-page}.

Although the literature is consistent about the need of document context in human evaluation of MT, little research has been done on the impact of experimental design and user interfaces on annotator productivity and the reliability of assessments in this context. 
The existing research on experimental designs for machine translation evaluation focuses on contrasting direct assessments with pairwise rankings \cite{novikova-etal-2018-rankme,sakaguchi-van-durme-2018-efficient} and not on the optimal presentation of the document-level information.
However, even the simple UI design decision of aligning document translations on the sentence level impacts efficiency of some evaluators \cite{popovic-2020-informative}.
With this work, we want to promote that direction of research.

\section{Document-level human evaluation campaigns at WMT}
\label{sec:interfaces}

During the WMT evaluation campaigns of 2019 and 2020, segment and document-level assessments of document translations were collected, but using different methods and thus user interfaces. Both were implemented in the Appraise evaluation framework \cite{federmann-2018-appraise} as a source-based direct assessment task \cite{graham-etal-2013-continuous,cettolo2017overview}, i.e.~all segments and entire documents were judged on a continuous scale between 0 and 100 by bilingual annotators.

\subsection{The WMT19 interface}

At WMT19, the evaluation of a translated document consisted of two parts: first, an evaluator would rate all individual segments in a document translated by one MT system, one by one, in the order they appear in the document, followed by assigning a single score to the whole document. Evaluators would be presented with the translation of a single segment (a source sentence and its translation) per screen, or the translation of the entire document.
Figures~\ref{fig:wmt19seg} and~\ref{fig:wmt19doc} depict segment-level and document-level portions of the interface, respectively.

This method was a simple document-level extension of the purely segment-level evaluations hosted during the previous editions of the WMT evaluation campaigns and did not require significant changes to the UI. A consequence of this approach was limited inter-sentential context as discussed by \newcite{toral-2020-reassessing}, since evaluators could not revisit the previously rated segments nor see subsequent ones. A rating decision could not be corrected in the light of the later-revealed context.

\subsection{The WMT20 interface}

At WMT20, both segment-level and document-level evaluations were performed on one screen. An evaluator would be presented with a translation of the entire document produced by one MT system. The document and its translation would be placed on a single vertically scrollable screen in two columns with source sentences on the left and their machine-translated counterparts on the right, aligned at segment-level. Figure~\ref{fig:wmt20} depicts a screenshot of this interface.

In the default scenario, the evaluator would be rating individual segments sequentially and, after rating all segments, on the same screen, the evaluator would rate the translation of the entire document at the bottom of the screen. Evaluators could, however, re-visit and update scores of previously rated segments at any time while still assessing the given document. They could also expand all sliders individually or in full, allowing them to take in all previously assigned scores.

\section{Human assessment data}
\label{sec:data}

\begin{table}[t]\centering
\small
\begin{tabular}{clrr}
\toprule
    & Statistic & WMT19 & WMT20 \\
\midrule
    All    & Languages          & cs, de, fi, gu & cs, de, iu, jp \\
           &                    & kk, lt, ru, zh & pl, ru, ta, zh \\
           & Annotators         & 1,271     & 1,213     \\
           & Seg. judgements    & 207,916   & 186,813   \\
           & Doc. judgements    & 12,907    & 13,790    \\
\midrule
    L4   & Languages          & cs, de, ru, zh & cs, de, ru, zh \\
           & Annotators         & 779       & 746       \\
           & Seg. judgements    & 127,178   & 115,571   \\
           & Doc. judgements    & 7,894     & 10,019    \\
\bottomrule
\end{tabular}
\caption{Statistics of data from the WMT19 and WMT20 campaigns, including languages, the total number of annotators and collected segment-level and document-level scores, after excluding documents with quality control items.}
\label{tab:data}
\end{table}

\begin{table*}[t]\centering
\small
\begin{tabular}{lrrrrr}
\toprule
 & \multicolumn{2}{c}{WMT19} & \multicolumn{2}{c}{WMT20} & \\
 & Avg. & Std. & Avg. & Std. & $\Delta$ (\%) \\
\midrule
Annotation time for a task (200 seg.)          &  1:06:08 &    $\pm$ 21:47 &  1:51:09 &    $\pm$ 51:12 & +68.1  \\
\midrule                                                              
Total time for documents $<$10 seg. &    02:02 &    $\pm$ 01:00 &    02:48 &    $\pm$ 01:44 & +37.1  \\
Total time for documents $>$20 seg. &    06:54 &    $\pm$ 02:48 &    12:01 &    $\pm$ 07:53 & +74.0  \\
\midrule                             
Time for 1st half of documents      &    02:06 &    $\pm$ 01:09 &    02:44 &    $\pm$ 02:05 & +30.5 \\
Time for 2nd half of documents      &    01:50 &    $\pm$ 00:58 &    01:53 &    $\pm$ 01:22 & +2.4  \\
Time for first 3 seg. in documents  &    00:52 &    $\pm$ 00:24 &    01:26 &    $\pm$ 01:02 & +66.3  \\
Time for last 3 seg. in documents   &    00:42 &    $\pm$ 00:18 &    00:51 &    $\pm$ 00:30 & +20.4  \\
\midrule                             
Time for single segment score      &    00:16 &    $\pm$ 00:06 &    00:24 &    $\pm$ 00:13 & +47.4  \\
Time for single document score     &    00:12 &    $\pm$ 00:09 &    00:06 &    $\pm$ 00:04 & -42.7 \\
\bottomrule
\end{tabular}
\caption{Average annotation times with standard deviations for tasks, documents, parts of documents and segments in the \textit{(hours):minutes:seconds} format.}
\label{tab:time}
\end{table*}

In our experiments, we utilize the human assessment data collected at the WMT19 and WMT20 evaluation campaigns. We limit the data to out-of-English language pairs as the into-English evaluation at WMT20 was done using the WMT19 method of reference-based DA and assessed by crowd workers instead of translators and researchers. Each annotator account provided 200 segment-level scores, and a number of document-level scores depending on the length of documents in the annotator's sample. From our analysis, we exclude all documents that contain one or more quality control segments, which constitute about 12\% of all segments.\footnote{Please refer to \newcite{barrault-etal-2020-findings} for more details on the quality control methods used at WMT.}

We use similar amounts of assessments from both campaigns, as seen in Table~\ref{tab:data}: WMT19 provided 208K segment and 13K document ratings, while 187K and 14K were collected for WMT20, respectively. We either compare data collected for all eight languages in each campaign or only subsets from four languages that were present in both years, i.e. Czech, German, Russian, and Chinese, minimizing differentiation factors between the data.
Note that the WMT19 and WMT20 assessment data concern disjoint sets of segments as \emph{different} test sets and MT systems were evaluated in both campaigns. We are interested in general patterns in the data at a larger scale, so we do not perceive this as an issue, but are aware of the fact in our conclusions. In a more ideal situation, we would have been able to perform A/B testing of different interfaces at the same campaign, but this was not an available option during the actual campaigns.

\section{Experiments on WMT data}
\label{sec:experiments}

We aim at comparing the WMT19 and WMT20 interfaces for segment and document-level human assessments of MT outputs by analyzing the data that has been collected using both methods. We analyze annotation times, compare correlations of document and averaged segment ratings, and examine the inter-annotator agreement.

\subsection{Annotation times}

We analyze annotation times to examine if and to what extent document context is used by annotators if it is available to them during assessment of individual segments.

In both interfaces, two timestamps were collected for each segment or document. In WMT19, timestamps were recorded when a new page opened and when an annotator submitted a score. In the WMT20 document-level interface timestamps were recorded when a segment was (automatically or manually) expanded and when a score was submitted. Note that in the WMT20 campaign, annotators see all segments during the assessment of the document and can read ahead even before the first timestamp is collected. This could make the collected annotation times for WMT20 slightly less reliable.

We report annotation time statistics only for evaluators who completed their task consisting of 200 segments (74\% of evaluators at WMT19 and 84\% at WMT20). 
Very quickly annotated items indicate users who potentially gamed the task and assigned random scores. Items that took an excessive amount of time were likely interrupted with unrelated activity or otherwise idle. In order to account for these situations, we remove data points with values smaller than the 10th percentile or larger than the 90th percentile. The results are shown in Table~\ref{tab:time}.

Our observations are as follows:
\begin{itemize}
    \item Providing the full document context increases the total annotation time per task by 68\% on average. This suggests that annotators do read the context and use it during assessments. Significantly increased annotation time raises the question about cost efficiency of the document-centric evaluations. 
    
    \item The more context is available, the more time annotators spend on studying it: during WMT20, annotators spent 74\% more time on documents with 20 or more segments than on documents of similar length during WMT19, whereas the per-document annotation time for shorter documents with 10 or fewer segments increased by only 37\%. 
    
    \item Comparing the average annotation times for segments from the beginning of the document with those farther into the documents, we can see that with the WMT20 interface annotators significantly increase the pace of annotation throughout the assessment of segments in a document. this is much less prominent for WMT19, which suggests that annotators do read the context ahead before making assessments \cite{castilho-2020-page} and that they can memorize and make better use of the preceding context if it is available to them at all time.
\end{itemize}

As described in Section~\ref{sec:interfaces}, the new interface allowed annotators to revise any segment score in a document before submitting the document score. We found that annotators did not use this feature often, and only 1.9\% segment-level scores were revised, which resulted in 9.0\% documents with one or more revised scores.

These observations suggest that annotators do make use of the available context and spend additional time studying it. Whether using that context results in more reliable quality assessments at segment and document level remains however unanswered.

\subsection{Correlation of document and segment-level judgements}

\begin{table}[t]
\small
\begin{subtable}{.47\textwidth}
\centering
\begin{tabular}{p{2.7cm}rrr}
\toprule
Aggregation  & WMT19 & WMT20   & $\Delta$ \\
\midrule
Avg. seg. score        &  0.907  &  0.923  &  0.016 \\
Min. seg. score        &  0.723  &  0.736  &  0.013 \\
Max. seg. score        &  0.584  &  0.628  &  0.044 \\
Avg. of first 5 seg.   &  0.845  &  0.861  &  0.015 \\
Avg. of last 5 seg.    &  0.883  &  0.899  &  0.016 \\
\midrule
Avg.~short~doc.~1\textsuperscript{st}~half & 0.827 & 0.841           & --\\
Avg.~short~doc.~2\textsuperscript{nd}~half & 0.887 & 0.901           & --\\
Avg.~long~doc.~1\textsuperscript{st}~half  & 0.868 & \textbf{0.893}  & --\\
Avg.~long~doc.~2\textsuperscript{nd}~half  & 0.894 & \textbf{0.909}  & --\\
\bottomrule
\end{tabular}
\caption{\label{tab:corrall}All languages}
\vspace{1em}
\end{subtable}
\begin{subtable}{.47\textwidth}
\centering
\begin{tabular}{p{2.7cm}rrr}
\toprule
Aggregation  & WMT19 & WMT20   & $\Delta$ \\
\midrule
Avg. seg. score        &  0.862  &  0.919 & 0.057 \\
Min. seg. score        &  0.658  &  0.761 & 0.103 \\
Max. seg. score        &  0.520  &  0.648 & 0.128 \\
Avg. of first 5 seg.   &  0.786  &  0.865 & 0.078 \\
Avg. of last 5 seg.    &  0.830  &  0.903 & 0.073 \\
\bottomrule
\end{tabular}
\caption{4 common languages (cs, de, ru, zh)}
\end{subtable}

\caption{Pearson correlations between document-level scores and different aggregations of segment-level scores: average, minimum, maximum, average of first or last 5 segments in the document.}
\label{tab:corr}
\end{table}

We measure the similarity between document-level scores and aggregated segment-level scores using different statistics, for example an average, from the same documents. We use the Pearson coefficient as the correlation measure \cite{freedman2007statistics}. We hypothesize that an increased correlation may be contributed to an improved capability of the user interface for reliable assessment of document translations by annotators.

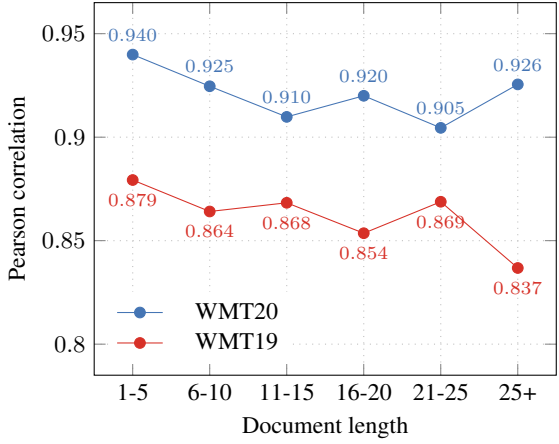
\begin{figure}[t]
    \centering
\begin{tikzpicture}
    \small
    \begin{axis}[
            height=0.2\textheight,
            width=0.38\textwidth,
            scale only axis,
            enlarge y limits,
            ymajorgrids, xmajorgrids,
            major grid style={dotted},
            xtick={1,2,3,4,5,6},
            legend cell align=left,
            legend style={column sep=10pt},
            every node near coord/.append style={anchor=south, font=\scriptsize, xshift=0pt, /pgf/number format/.cd, fixed, fixed zerofill, precision=3, /tikz/.cd},
            xticklabels={1-5, 6-10, 11-15, 16-20, 21-25, 25+},
            ymax=0.95,
            ymin=0.8,
            xmin=0.5,
            xmax=6.5,
            legend pos=south west, legend style={draw=none,font=\footnotesize},
            xticklabel style={font=\footnotesize},
            ylabel={Pearson correlation},
            xlabel={Document length},
        ]

        \addplot+[solid, nodes near coords, 
            every node near coord/.append style={anchor=south, yshift=2pt},
            mark=*, mark options={solid}, myblue]
        table[x index=0, y=wmt20] {corrvslen};

        \addplot+[nodes near coords, 
            every node near coord/.append style={anchor=north, yshift=-2pt},
            solid, mark=*,mark options={solid}, myred]
        table[x index=0, y=wmt19] {corrvslen};
        
        \legend{WMT20,WMT19}
    \end{axis}
\end{tikzpicture}
    \caption{Pearson correlations between document-level and the average of segment-level scores in relation to the number of segments in the document (4 common languages).}
    \label{fig:corrlen}
\end{figure}

Our main results are presented in Table~\ref{tab:corr} and Figure~\ref{fig:corrlen}.
We excluded all documents that contained one or more segments used for quality control (26\% and 22\% for WMT19 and WMT20, respectively) before computing the correlation statistics.
We did not exclude scores from users who did not pass the quality control as this is not practiced by the WMT organizers when computing human rankings of MT systems for out-of-English languages. These users contributed only a small fraction of the data and excluding their scores does not meaningfully change the results. The scores were not standardized prior to computation. 

We observe the following effects of the WMT20 interface compared to the WMT19 interface:
\begin{itemize}
    \item We can see consistently higher correlations between document-level scores and all tested aggregations of segment-level scores for WMT20. This effect is even more prominent on the four common language pairs used in both campaigns.
    
    \item Document-level scores show the highest correlation with the averaged segment-level scores. The very high correlation of 0.92 indicates that the average of segment ratings from a document might be used as a reasonable approximation of the final document ratings in the document-centric evaluation. This might justify dropping the final document score from the assessment.
    
    \item The lowest segment score in documents correlates better with the overall document score than the highest segment score (\textit{Min.~seg.} vs \textit{Max.~seg.}). Intuitively, badly translated segments may impact the overall perception of the document quality more than higher-quality segment translations, or this could be attributed to the fact that shorter sentences are more likely to be translated correctly, but annotators may not see them as contributive to the overall document translation quality as longer sentences.
    
    \item Regardless of the user interface, segments from the end of a document influence assessment of the entire document more than segments from the beginning of the document (\textit{Avg.~of first 5} vs \textit{Avg.~of last 5}). From this, we do not observe that showing segments sequentially penalizes the very first segments in the document in contributing to the overall document score. However, the comparison of correlations for short and long documents (up to 10 segments, or more than 20 segments; bottom part of Table~\ref{tab:corrall}) reveals that WMT20 seems to improve the contribution of early segments to the document score for long documents.
    
    \item In Figure~\ref{fig:corrlen}, we computed correlations for averaged segment-level scores in relation to the number of segments in documents. Interestingly, for WMT20, the correlation increases for the longest documents (more than 25 segments).
\end{itemize}

The same trends are observed if Spearman's or Kendall's rank correlation coefficients are used instead of Pearson's correlation coefficient.

\subsection{Inter-annotator agreement}
\label{ssec:iaa}

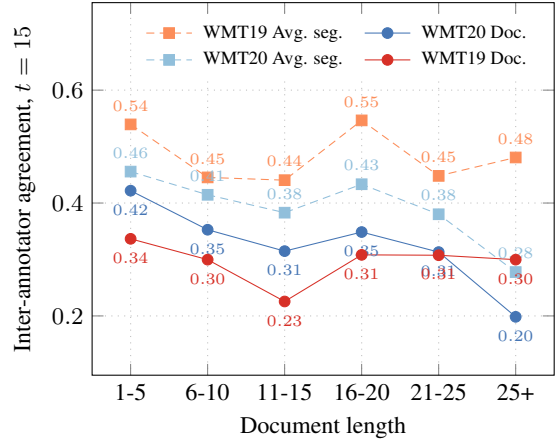
\begin{figure}[t]
    \centering
\begin{tikzpicture}
    \small
    \begin{axis}[
            height=0.2\textheight,
            width=0.38\textwidth,
            scale only axis,
            enlarge y limits,
            ymajorgrids, xmajorgrids,
            major grid style={dotted},
            xtick={1,2,3,4,5,6},
            legend cell align=left,
            legend style={column sep=10pt},
            every node near coord/.append style={anchor=south, font=\tiny, xshift=0pt, /pgf/number format/.cd, fixed, fixed zerofill, precision=2, /tikz/.cd},
            xticklabels={1-5, 6-10, 11-15, 16-20, 21-25, 25+},
            ymax=0.7,  
            ymin=0.15,
            xmin=0.5,
            xmax=6.5,
            legend columns=2, 
            legend pos=north east, legend style={draw=none,font=\scriptsize,column sep=2pt,/tikz/column 2/.style={column sep=6pt,}},
            xticklabel style={font=\footnotesize},
            ylabel={Inter-annotator agreement, $t=15$},
            xlabel={Document length},
        ]

        \addplot+[densely dashed, mark=square*,mark options={solid}, mylightred,
            nodes near coords,  every node near coord/.append style={anchor=south, yshift=2pt}
            ] table[x index=0, y=wmt19seg] {iaavslen};
            
        \addplot+[solid, mark=*, mark options={solid}, myblue,
            nodes near coords, every node near coord/.append style={anchor=north, yshift=-2pt}
            ] table[x index=0, y=wmt20doc] {iaavslen};
            
        \addplot+[densely dashed, mark=square*, mark options={solid}, mylightblue,
            nodes near coords, every node near coord/.append style={anchor=south, yshift=2pt}
            ] table[x index=0, y=wmt20seg] {iaavslen};

        \addplot+[solid, mark=*, mark options={solid}, myred,
            nodes near coords,  every node near coord/.append style={anchor=north, yshift=-2pt}
            ] table[x index=0, y=wmt19doc] {iaavslen};
        
        \legend{WMT19 Avg. seg.,WMT20 Doc.,WMT20 Avg. seg.,WMT19 Doc.}
    \end{axis}
\end{tikzpicture}
    \caption{Inter-annotator agreements (Cohen's kappa, $t=15$) for document-level and averaged segment-level scores in relation to the number of segments in the document (4 common languages).}
    \label{fig:iaalen}
\end{figure}
\begin{table*}[!ht]
\small
\setlength{\tabcolsep}{2pt}
\begin{subtable}{.5\textwidth} \centering
\begin{tabular}{crrR{1cm}|rrR{1cm}}
         & \multicolumn{3}{c}{Doc. scores} & \multicolumn{3}{c}{Avg. seg. scores} \\
\toprule
$t$      & WMT19 & WMT20 & $\Delta$         & WMT19 & WMT20 & $\Delta $\\
\midrule
5        & 0.110 & 0.118   & 0.007        & 0.148 & 0.132   & -0.016\\
10       & 0.195 & 0.215   & 0.020        & 0.290 & 0.254   & -0.036\\
15       & 0.280 & 0.333   & 0.053        & 0.433 & 0.390   & -0.044\\
20       & 0.378 & 0.443   & 0.065        & 0.560 & 0.514   & -0.046\\
25       & 0.481 & 0.554   & 0.073        & 0.669 & 0.634   & -0.035\\
30       & 0.559 & 0.639   & 0.080        & 0.760 & 0.737   & -0.023\\
\bottomrule
\addlinespace[.4em]
\multicolumn{4}{r}{Documents}               & 12,907 & 13,790 \\
\multicolumn{4}{r}{Distinct documents}      & 10,132 & 7,020  \\
\multicolumn{4}{r}{With multiple judgements}& 26.2\% & 66.1\% \\
\end{tabular}
\caption{All languages}
\end{subtable}
\hfil
\begin{subtable}{.5\textwidth} \centering
\begin{tabular}{crrR{1cm}|rrR{1cm}}
         & \multicolumn{3}{c}{Doc. scores} & \multicolumn{3}{c}{Avg. seg. scores} \\
\toprule
$t$      & WMT19 & WMT20 & $\Delta$         & WMT19 & WMT20 & $\Delta $\\
\midrule
5        & 0.115 & 0.124 & 0.009          & 0.182 & 0.144 &-0.039\\
10       & 0.202 & 0.226 & 0.024          & 0.329 & 0.272 &-0.057\\
15       & 0.302 & 0.343 & 0.040          & 0.481 & 0.406 &-0.075\\
20       & 0.403 & 0.456 & 0.053          & 0.637 & 0.536 &-0.101\\
25       & 0.509 & 0.569 & 0.059          & 0.756 & 0.657 &-0.100\\
30       & 0.580 & 0.648 & 0.068          & 0.851 & 0.753 &-0.098\\
\bottomrule
\addlinespace[.3em]
\multicolumn{4}{r}{Documents}               &  7,894 & 10,019 \\
\multicolumn{4}{r}{Distinct documents}      &  6,376 & 4,798  \\
\multicolumn{4}{r}{With multiple judgements}& 23.0\% & 74.3\% \\
\end{tabular}
\caption{4 common languages (cs, de, ru, zh)}
\end{subtable}

\caption{Inter-annotator agreement (Cohen's kappa) on document-level scores and averaged segment-level scores for different tolerances $t$, i.e. two scores are assumed equal if they differ no more than $t$.}
\label{tab:iaa}
\end{table*}

We compute annotator agreement as a measure of reliability between annotators with Cohen's kappa coefficient \cite{cohen1960}

$$
\kappa = \frac{P_a - P_e}{1 - P_e},
$$

\noindent
where $P_a$ is the observed proportion of times that two annotators agree, and $P_e$ is the expected mean proportion of agreement due to chance. Values of $\kappa$ close to 0 are interpreted as no agreement and $\kappa$ is equal to 1 if there is perfect agreement. 

$P_a$ is computed from pairwise comparisons of all documents that have been annotated by two or more annotators by counting the proportion of times that two annotators agree on the score.\footnote{If a document is annotated by more than two annotators, pairwise comparisons between all annotators are counted.} It is assumed that two annotators agree if their assigned scores $s_i$ and $s_j$ differ no more than a predefined tolerance $t$, i.e. $|s_i-s_j| \leq t$.

\begin{table}[t]\centering
\small
\begin{tabular}{lrrrrrr}
\toprule
$t$   &     5 &    10 &    15 &    20 &    25 &    30 \\
\midrule
$P_e$ & 0.107 & 0.199 & 0.286 & 0.368 & 0.445 & 0.517 \\
\bottomrule
\end{tabular}
\caption{Examples of $P_e$ for different tolerances $t$.}
\label{tab:pe}
\end{table}

$P_e$ is constant for a given $t$ and computed as the sum of probabilities of randomly assigning a score within the tolerance $t$ (inclusive) over all possible scores from 1 to 100, i.e.:

$$
P_e = \sum_{i \in [1,100]} \frac{\textit{min}(i+t, 100) - \textit{max}(i-t, 0) + 1}{100^2}.
$$

\noindent
Examples of $P_e$ for different $t$ are presented in Table~\ref{tab:pe}.

We compute inter-annotator agreement (IAA) for $t={5, 10, 15, 20, 25, 30}$, and compare agreement for document-level and averaged segment-level scores, presenting the results in Table~\ref{tab:iaa}.
Since there are very few annotators who have annotated the same documents more than once, we do not compute document-level intra-annotator agreement.

Here, our main observations are as follows:
\begin{itemize}
    \item Obviously, the larger the tolerance $t$, the higher the agreement.
    Because the average difference of document-level and segment-level scores for documents assessed multiple times is between 15.0 and 19.6 (not shown in the table), we can assume that a $t$ value of 15 or 20 is the most reasonable. In this case, the inter-annotator agreement is fair or sometimes moderate according to the recommended interpretation scale proposed by \newcite{Landis77}. 
    
    \item For both methods, agreement for document-level scores is lower than for segment-level scores.
    This confirms the finding of \newcite{castilho-2020-page} that document-level evaluation efforts where annotators assign one score per document leads to lower levels of inter-annotator agreement for adequacy when compared to segment-level evaluation. In contrary to that work, our analysis is done at a much larger scale and for multiple language pairs.
    
    \item Inter-annotator agreement of document-level scores is higher for WMT20 than for WMT19 (4th column). Interestingly, the opposite is true for averaged segment-level scores (7th column), and it is even more prominent for the subset of four common languages. We will discuss this some more in Section~\ref{sec:discussion}.
    
    \item As shown in Figure~\ref{fig:iaalen}, inter-annotator agreement decreases with increasing document length for WMT20, but it flattens for the longest documents in the case of WMT19. 
\end{itemize}

In Appendix~\ref{sec:appendix} we provide inter-annotator agreement results computed with the Krippendorff's alpha coefficient \cite{krippendorff} for reference.

\section{Discussion}
\label{sec:discussion}

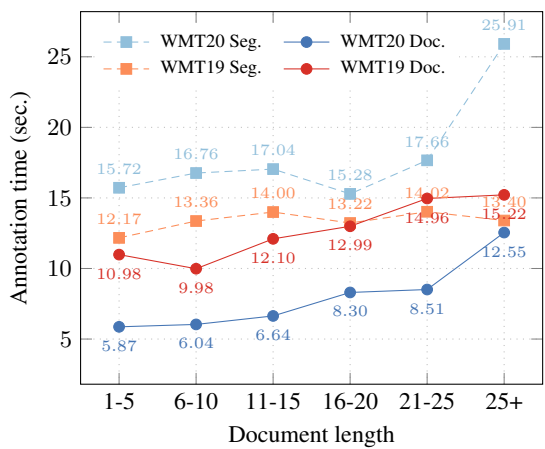
\begin{figure}[t]
    \centering
\begin{tikzpicture}
    \small
    \begin{axis}[
            height=0.2\textheight,
            width=0.38\textwidth,
            scale only axis,
            enlarge y limits,
            ymajorgrids, xmajorgrids,
            major grid style={dotted},
            xtick={1,2,3,4,5,6},
            legend cell align=left,
            legend style={column sep=10pt},
            every node near coord/.append style={anchor=south, font=\tiny, xshift=0pt, /pgf/number format/.cd, fixed, fixed zerofill, precision=2, /tikz/.cd},
            xticklabels={1-5, 6-10, 11-15, 16-20, 21-25, 25+},
            ymax=26,  
            ymin=4,
            xmin=0.5,
            xmax=6.5,
            legend columns=2, 
            legend pos=north west, legend style={draw=none,font=\scriptsize,column sep=2pt,/tikz/column 2/.style={column sep=6pt,}},
            xticklabel style={font=\footnotesize},
            ylabel={Annotation time (sec.)},
            xlabel={Document length},
        ]

        \addplot+[densely dashed, mark=square*, mark options={solid}, mylightblue,
            nodes near coords, every node near coord/.append style={anchor=south, yshift=2pt}
            ] table[x index=0, y=wmt20seg] {timevslen};
            
        \addplot+[solid, mark=*, mark options={solid}, myblue,
            nodes near coords, every node near coord/.append style={anchor=north, yshift=-2pt}
            ] table[x index=0, y=wmt20doc] {timevslen};
            
        \addplot+[densely dashed, mark=square*,mark options={solid}, mylightred,
            nodes near coords,  every node near coord/.append style={anchor=south, yshift=2pt}
            ] table[x index=0, y=wmt19seg] {timevslen};

        \addplot+[solid, mark=*, mark options={solid}, myred,
            nodes near coords,  every node near coord/.append style={anchor=north, yshift=-2pt}
            ] table[x index=0, y=wmt19doc] {timevslen};
        
        \legend{WMT20 Seg.,WMT20 Doc.,WMT19 Seg.,WMT19 Doc.}
    \end{axis}
\end{tikzpicture}
    \caption{Annotation times (sec.) for single segment or document score in relation to the number of segments in the document (all languages).}
    \label{fig:timelen}
\end{figure}

In the presented experiments, we have observed interesting differences in correlation and inter-annotator agreement for long documents. In WMT19, for long documents, the correlation between segment-level scores and document-level scores significantly decreases, while IAA flattens out and eventually ends up being higher than for WMT20. We think this might be an effect of cognitive overload when annotators are presented with long document translation text pairs without visual help in the form of sentence alignment and similar hints.\footnote{See the example on Figure~\ref{fig:wmt19doc} consisting only of 6 segments. A thoughtful evaluation of an article with 20 or more segments would appear even more challenging.} A large wall of text might discourage annotators and they might fall back to assigning default or less diverse ``safe'' scores. Analyzing annotation times in relation to the document length, presented in Figure~\ref{fig:timelen} supports this explanation. The average time of document ratings flattens for documents longer than 20 segments for WMT19, while it increases for WMT20.

Another non-intuitive observation we have made is that the inter-annotator agreement for averaged segment scores is higher in WMT19 than in WMT20. The agreement for document scores is, as expected, consistently higher for WMT20. If this is not solely attributed to the different data sets used in both campaigns, we would explain it by a tendency of annotators to assign higher scores if they cannot identify errors due to insufficient context \cite{laubli2020jair}, which may occur for WMT19 because of its limited inter-sentential context. Another explanation would be that the WMT20 interface presenting all sentences from the document at once, encourages annotators to assign more diversified scores across segments; this may then lower the agreement at segment level. However, we were not able to confirm this based on an analysis of histograms of segment scores and their standard deviations.

Our study is conducted post-hoc, i.e. we cannot test for scenarios that were not anticipated during the actual evaluation campaigns. 
A more conclusive interpretation would require A/B testing with the same sets of documents, translations and annotators used for both evaluation methods. Nevertheless, we think that the presented comparison of two WMT evaluation campaigns supports the assumption that the document-centric evaluation conducted during WMT20 produced more reliable document ratings. We believe this to be an important finding because higher quality of collected document assessments should help to avoid statistical issues arising from low statistical power as observed by \newcite{graham-etal-2020-statistical}.

\section{Summary}
\label{sec:summary}

In this work, we have compared two methods for document-level human evaluation of MT outputs through an analysis of the large-scale human assessment data from WMT evaluation campaigns, consisting of 8 different out-of-English language pairs. Our main findings are:
\begin{itemize}
    \item Showing the entire document can extend the annotation time of individual segments by as much as 68\% --- presumably because annotators make use of the available context during evaluation.
    \item Annotators rarely change their segment-level ratings even if this option is available to them. Nevertheless, in some instances they do.
    \item Annotators tend to rate documents more consistently with their segment ratings if the entire document context is available at all time.
    \item In the document-centric evaluation, document ratings can be approximated reasonably well by averaged segment level scores.
    \item Inter-annotator agreement for document ratings increases if segment level evaluation is made in the global context.
\end{itemize}

Our analysis suggests that not only the entire document context is needed for reliable human evaluation of news translations, as recent studies have shown, but that the method in which the context is presented to evaluators is also important for collecting good-quality segment and document-level assessments. 
We conclude that the WMT20 method produces more reliable ratings, and thus can be adopted for future editions of the WMT document-level human evaluation campaigns for all languages. 

In future work, we plan to strengthen our findings by comparing the WMT19 and WMT20 methods in A/B testing with common sets of documents, translations and annotators for both settings.

\bibliography{interfaces}
\bibliographystyle{acl_natbib}

\appendix
\section{Appendix}
\label{sec:appendix}

\begin{figure}[!ht]
    \centering
\begin{tikzpicture}
    \small
    \begin{axis}[
            height=0.2\textheight,
            width=0.38\textwidth,
            scale only axis,
            enlarge y limits,
            ymajorgrids, xmajorgrids,
            major grid style={dotted},
            xtick={1,2,3,4,5,6},
            legend cell align=left,
            legend style={column sep=10pt},
            every node near coord/.append style={anchor=south, font=\tiny, xshift=0pt, /pgf/number format/.cd, fixed, fixed zerofill, precision=2, /tikz/.cd},
            xticklabels={1-5, 6-10, 11-15, 16-20, 21-25, 25+},
            ymax=0.7,  
            ymin=0.15,
            xmin=0.5,
            xmax=6.5,
            legend columns=2, 
            legend pos=north east, legend style={draw=none,font=\scriptsize,column sep=2pt,/tikz/column 2/.style={column sep=6pt,}},
            xticklabel style={font=\footnotesize},
            ylabel={Inter-annotator agreement (interval scale)},
            xlabel={Document length},
        ]

        \addplot+[densely dashed, mark=square*,mark options={solid}, mylightred,
            nodes near coords,  every node near coord/.append style={anchor=south, yshift=2pt}
            ] table[x index=0, y=wmt19seg] {iaa2vslen};
            
        \addplot+[solid, mark=*, mark options={solid}, myblue,
            nodes near coords, every node near coord/.append style={anchor=north, yshift=-2pt}
            ] table[x index=0, y=wmt20doc] {iaa2vslen};
            
        \addplot+[densely dashed, mark=square*, mark options={solid}, mylightblue,
            nodes near coords, every node near coord/.append style={anchor=south, yshift=2pt}
            ] table[x index=0, y=wmt20seg] {iaa2vslen};

        \addplot+[solid, mark=*, mark options={solid}, myred,
            nodes near coords,  every node near coord/.append style={anchor=north, yshift=-2pt}
            ] table[x index=0, y=wmt19doc] {iaa2vslen};
        
        \legend{WMT19 Avg. seg.,WMT20 Doc.,WMT20 Avg. seg.,WMT19 Doc.}
    \end{axis}
\end{tikzpicture}
    \caption{Inter-annotator agreements (Krippendorff's alpha, interval metric) for document-level and averaged segment-level scores in relation to the number of segments in the document (4 common languages).}
    \label{fig:krippiaalen}
    \vspace{\textheight} 
\end{figure}
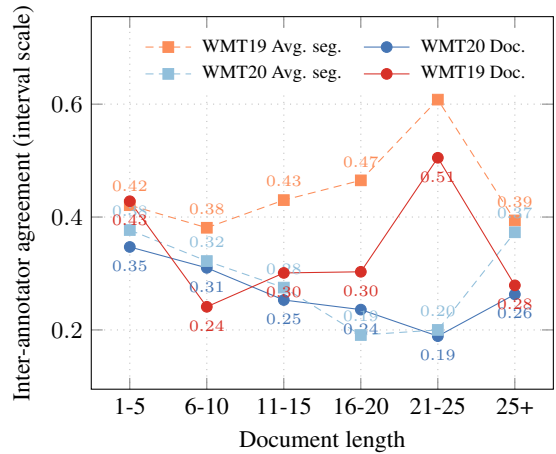

Table~\ref{tab:krippiaa} and Figure~\ref{fig:krippiaalen} provide inter-annotator agreement for document-level and averaged segment-level scores in the form of  Krippendorff's alpha coefficient \cite{krippendorff} for 4 common languages from WMT19 and WMT20.
We present coefficients computed with interval and ratio metrics, and for a direct comparison with the results presented in Section~\ref{ssec:iaa}, with the nominal metric with different tolerances $t$, i.e. two scores are assumed equal if they differ no more than $t$.

Krippendorff's alpha coefficients computed using the interval or ratio metrics do not show the higher agreement on document ratings for WMT20 compared to WMT19 that has been observed with Cohen's Kappa, but the difference is again smaller than for averaged segment ratings. Coefficients computed using the nominal metric with tolerance thresholds align with the inter-annotator agreement results obtained with the other statistic measure.

\begin{table}[!h]
\footnotesize
\setlength{\tabcolsep}{1.5pt}
\centering
\begin{tabular}{p{.6cm}rrR{1cm}|rrR{1cm}}
         & \multicolumn{3}{c}{Doc. scores} & \multicolumn{3}{c}{Avg. seg. scores} \\
\toprule
$t$       & WMT19 & WMT20 & $\Delta$         & WMT19 & WMT20 & $\Delta $\\
\midrule
Inter.    & 0.340 & 0.282 &-0.058& 0.465 & 0.297 &-0.168\\
Ratio     & 0.294 & 0.246 &-0.048& 0.461 & 0.277 &-0.184\\
\midrule
  5       & 0.030 & 0.046 & 0.016& 0.060 & 0.053 &-0.007\\
  10      & 0.061 & 0.077 & 0.016& 0.103 & 0.085 &-0.018\\
  15      & 0.100 & 0.130 & 0.030& 0.194 & 0.138 &-0.056\\
  20      & 0.153 & 0.188 & 0.035& 0.329 & 0.202 &-0.127\\
  25      & 0.237 & 0.258 & 0.021& 0.462 & 0.290 &-0.172\\
  30      & 0.286 & 0.311 & 0.025& 0.612 & 0.370 &-0.242\\
\bottomrule
\end{tabular}

\caption{Inter-annotator agreement (Krippendorff's alpha) on document-level and averaged segment-level scores for different metrics (4 common languages).}
\label{tab:krippiaa}
\end{table}

\end{document}